\documentclass[letterpaper]{article}
\usepackage{flairs}
\usepackage{times}
\usepackage{helvet}
\usepackage{courier}
\usepackage{booktabs}
\usepackage{array}
\usepackage{siunitx} 
\usepackage{makecell}
\usepackage{tcolorbox}
\usepackage{multirow}
\usepackage{latexsym}
\usepackage{graphicx}
\usepackage{booktabs}
\usepackage{url}
\usepackage{amsmath}
\usepackage{todonotes}
\begin{document}
\title{Exploring the Potential for Large Language Models \\ to Demonstrate Rational Probabilistic Beliefs}
\author{Gabriel Freedman, Francesca Toni\\
Department of Computing, Imperial College London, UK \\
    \{gif22, ft\}@imperial.ac.uk}

\maketitle
\begin{abstract}
Advances in the general capabilities of large language models (LLMs) have led to their use for information retrieval, and as components in automated decision systems. A faithful representation of probabilistic reasoning in these models may be essential to ensure trustworthy, explainable and effective performance in these tasks. Despite previous work suggesting that LLMs can perform complex reasoning and well-calibrated uncertainty quantification, we find
that current versions of this class of model lack the ability to provide rational and coherent representations of probabilistic beliefs. To demonstrate this, we introduce a novel dataset of claims with indeterminate truth values and apply a number of well-established techniques for uncertainty quantification to measure the ability of LLM's to adhere to fundamental properties of probabilistic reasoning.
\end{abstract}


\section{Introduction}

\begin{figure} 
\begin{minipage}{\columnwidth}
  \centering
  \includegraphics[scale=0.266, trim=1cm 0 0 0, clip]{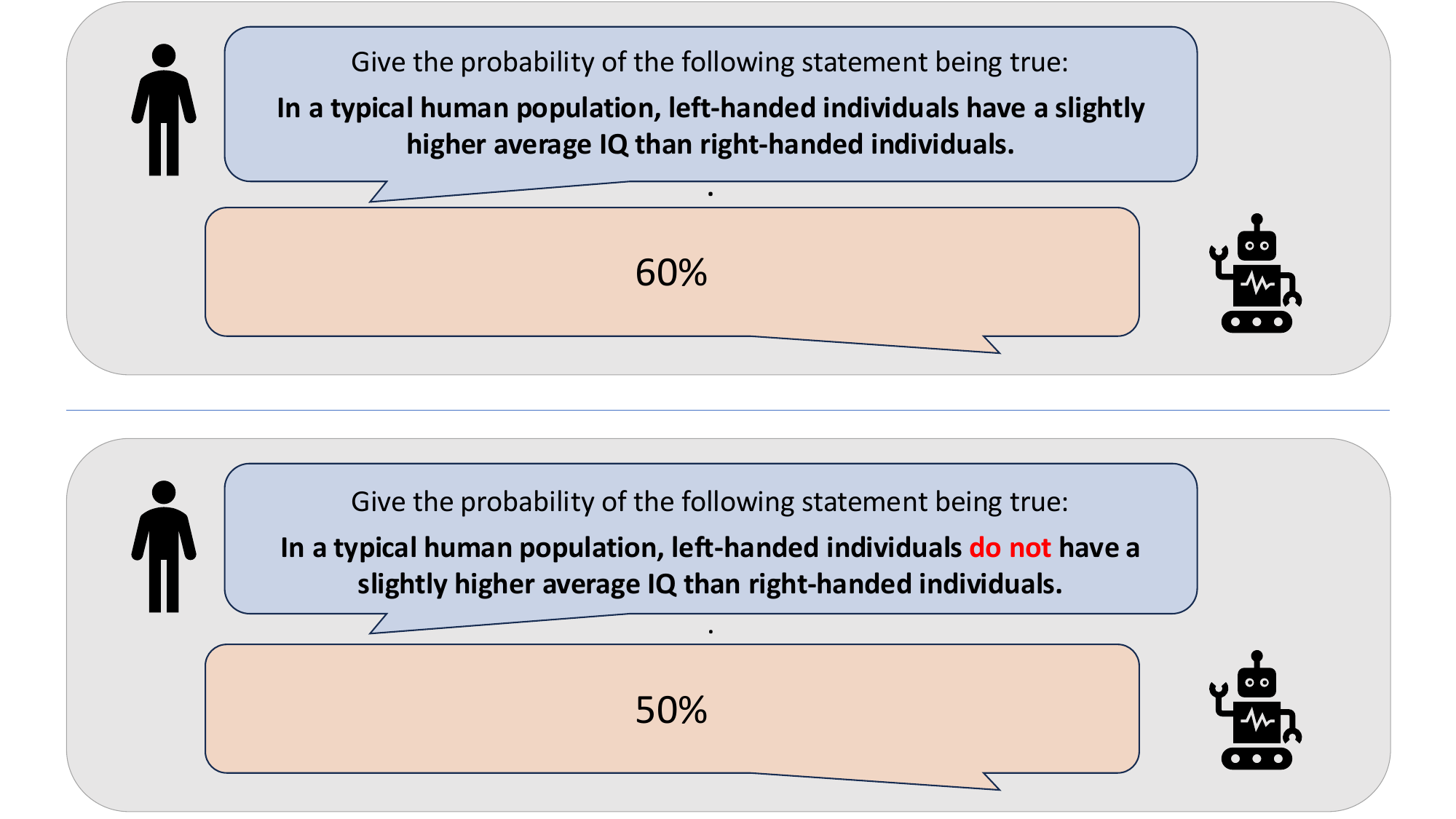} 
\end{minipage}

\vspace{3pt} 

\noindent 
\begin{minipage}{\columnwidth}
  \centering 
  \includegraphics[width=\linewidth]{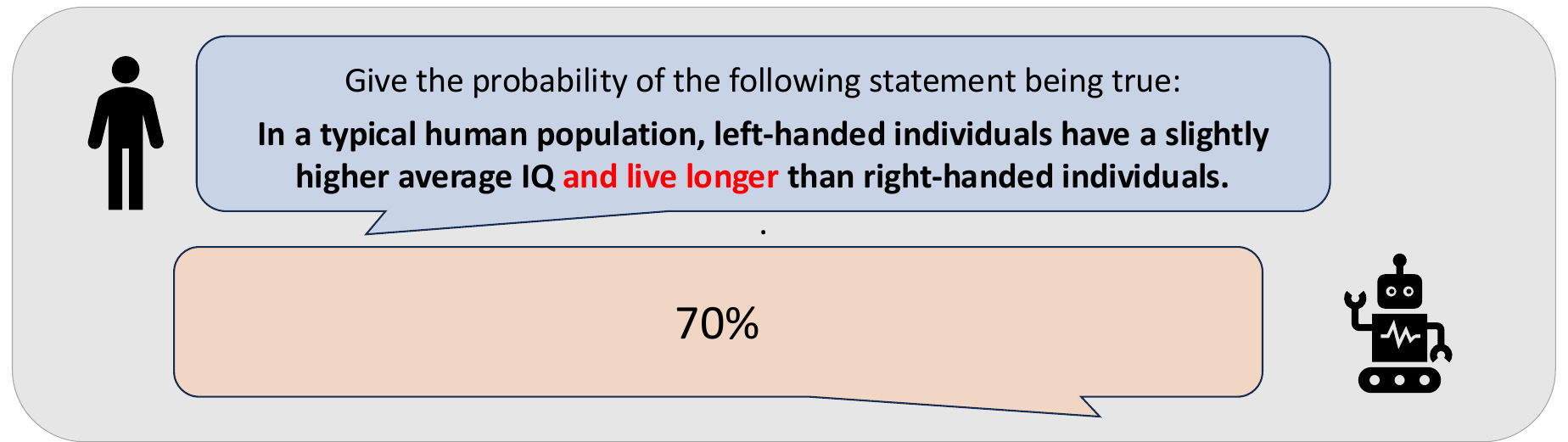} 
\end{minipage}

\vspace{4pt} 

\noindent 
\begin{minipage}{\columnwidth}
  \centering 
  \includegraphics[width=\linewidth]{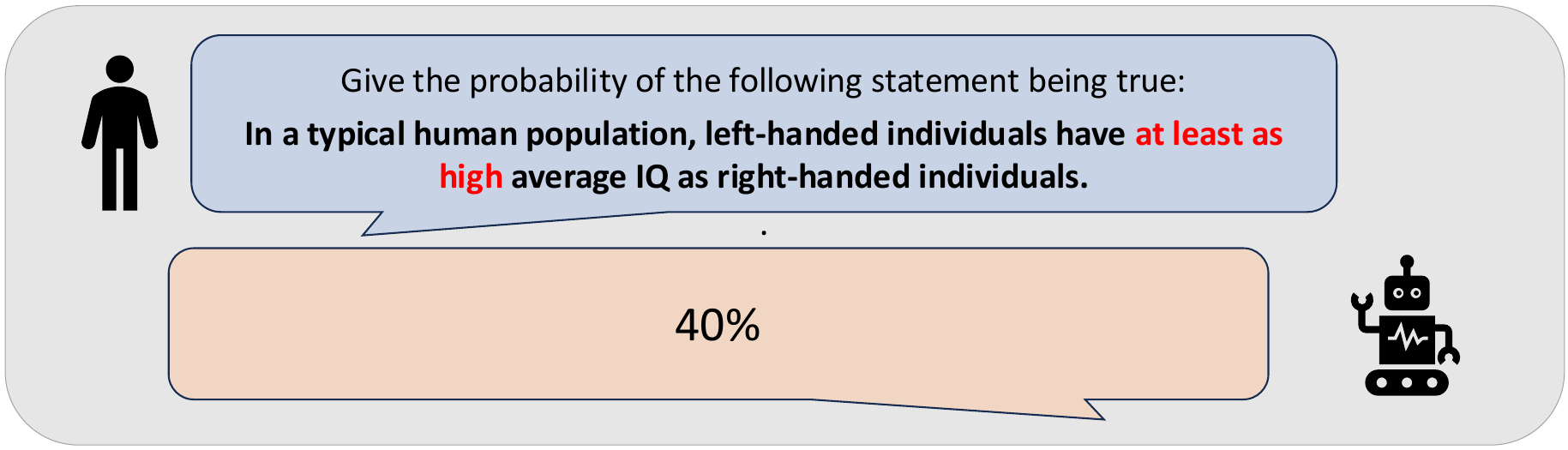} 
\end{minipage}

\caption
{Examples of LLMs violating three principles of probabilistic reasoning. The image above the line shows the original claim with the corresponding probability assignment by an LLM, and below the line the images demonstrate the violation of the principles of: \textit{complementarity} (top), \textit{specialisation} (middle) and \textit{generalisation} (bottom).}
\label{fig:examples}
\end{figure}

In order for an agent to be an effective probabilistic reasoner, it must not violate the axioms of probability \cite{Bas2019BasicCA}. If an agent 
violates 
any of these 
axioms, it implies that it lacks the capacity to perform robust probabilistic reasoning, including uncertainty quantification. 
Three of the most fundamental properties of probabilistic reasoning, derived as corollaries from Kolmogorov’s three axioms of probability theory \cite{kolmogorov1963theory}, are as follows. These properties clarify essential relationships between events, their complements, and their probabilities (for examples of each see Figure \ref{fig:examples}):

\begin{itemize}
    \item \textbf{Complementarity}: Consider any event \(A\) within the sample space \(\Omega\), where \(\Omega\) represents the set of all possible outcomes. Let \(A^c\) denote the complement of \(A\), defined as the set of all outcomes in \(\Omega\) that do not belong to \(A\). The property of complementarity states that:
    \[
    P(A) + P(A^c) = 1.
    \]
    This result arises directly from the axioms of normalisation (the probability of the entire sample space is 1) and finite additivity (the probability of the union of mutually exclusive events is the sum of their probabilities).

    \item \textbf{Monotonicity (Specialisation)}: For two events \(A\) and \(A'\) such that \(A' \subset A\) (i.e. $A'$ is more specific than $A$), the probability of \(A'\) is less than or equal to the probability of \(A\). Formally:
    \[
    P(A') < P(A).
    \]
    This reflects the principle that the probability of a more specific or refined event (a subset of another event) cannot exceed the probability of the broader event it is contained within.

    \item \textbf{Monotonicity (Generalisation)}: For two events \(A\) and \(A'\) such that \(A \subset A'\) (i.e. $A'$ is more general than $A$), the probability of \(A\) is less than or equal to the probability of \(A'\). Formally:
    \[
    P(A) < P(A').
    \]
    This is the inverse of the specialisation relation.
\end{itemize}

Generative LLMs have demonstrated impressive performance in many reasoning tasks - including tasks which they have not been specifically trained for \cite{Brown2020LanguageMA,Bubeck2023SparksOA}. This has led to the incorporation of LLMs into automated decision systems (ADSs) \cite{Zhang2023IntegratingAK,Ouyang2023AutoPlanAP,Wang2023RCAgentCR}.
In order for ADSs to be trustworthy and contestable \cite{Henin2021BeyondEJ,Lyons2021ConceptualisingC}, they should be accompanied by a faithful representation of their reasoning. In the 
majority of real-world settings, in order for this to be an effective representation, it would need to include probabilistic uncertainty estimates.

Unlike some previous work, we are attempting to measure something distinct from `subjective' uncertainty estimates \cite{Geng2023ASO}, which aim to assess the uncertainty intrinsic to the model. This is measured by comparing the uncertainty estimate with the veracity of the model's output. Instead, we are interested in what 
may be called `objective' uncertainty, as described in \cite{MIT2019}: \textit{``The objective probability of A
at time t is the subjective probability that a perfectly rational agent
would assign to A, if she had perfect information about the world at
times $\leq$ t and no information about the world at times \textgreater t.''}
. 
Thus, this is concerning the probability of a state of the world, regardless of the knowledge, or lack thereof, of the agent assigning the probability.  
This class of statements 
are at the core of academic disciplines and event forecasting. 

An example of a decision which requires `objective' uncertainty estimations is determining the diets of the first inhabitants of America. The main sources of uncertainty in this inference come from imprecision in the measurement of carbon isotope levels in bone samples, as well as in auxiliary evidence about the climate \cite{Booker2011AnEO}. Having a consistent and rational model of the uncertainties involved in this scenario allows for the effective integration of any new evidence or theories that come to light, and the possibility of overhauling 
existing conclusions.



In this paper we demonstrate that the current generation of state-of-the-art LLMs, both open and closed source, frequently violate the basic principles of probabilistic reasoning, even when applying techniques for enhancing the ability of LLMs to perform uncertainty quantification. This undermines a corpus of existing work that argues that LLMs exhibit sophisticated decision-making capabilities \cite{Bubeck2023SparksOA}, including the ability to effectively quantify uncertainty \cite{Lin2023GeneratingWC,tian2023just,Hou2023DecomposingUF}.

In Section~\ref{sec:dataset} we detail the construction of a novel dataset we use for evaluating LLMs' probabilistic beliefs. In Section~\ref{sec:tasks}  we describe our method for evaluating LLMs' adherence to the principles of complementarity and monotonicity and in Section~\ref{sec: uncertainty} we outline our methodologies for measuring uncertainty. In Section~\ref{sec:exp} we report the results of our experiments using these methods. In Section~\ref{sec:related} we briefly highlight relevant related work, and in the final Section we discuss the implications of our results and propose possible directions for future work.
Code and data can be found at \url{https://github.com/GIFRN/Rational-Probabilistic-Beliefs}.

\section{Novel Dataset}
\label{sec:dataset}

A contribution of this work is the construction of a novel, synthetically generated dataset of 517 claims with indeterminate truth values. Each claim is accompanied by its complement, as well as a more specific and more general version of the claim, amounting to 2068 samples in total. We name this the Rational Probabilistic Belief (RPB) dataset
.

The initial claims were generated by prompting GPT-4o to generate claims that had an \emph{n\%} probability of being true, with \emph{n} ranging between 10\% and 90\% in 10\% intervals. To promote variety in the types of claims, we generated three variations of the basic prompt: asking the LLM to pose as an expert, and asking for historical claims and scientific claims. 200 claims were generated using each version of the prompt, resulting in 800 total.

In order to further ensure sufficient diversity of claims, we filtered out similar claims. We did this with simple string matching, removing any claim that overlapped with any other claim in the dataset by at least 90\%. While this naive method does not capture semantically similar claims that are expressed differently, we note that the exact formulation of a statement sometimes has an impact on the uncertainty estimations of the LLMs, therefore making it desirable to have some samples where this is the case. We then manually remove any claim which is not making a valid statement that can be reasonably assigned a probability value. This leaves us with 517 base claims.

In order to generate the complementary, specialised and generalised versions of the claim, we also rely on GPT-4o. We use few-shot prompting with manually generated examples for each type of modification. We make sure the few-shot examples, especially for the specialised and generalised adaptions, are well chosen. This is because there is no set way to increase/decrease the specificity of a claim, and in order for it to create an appropriately challenging task, the newly generated claims must not be too distant (in a probabilistic sense) from the original. 

For instance, if we consider the example in Figure \ref{fig:examples}, a logically valid specialisation of the original claim is: `In a typical human population, left-handed individuals have a slightly higher average IQ and are more than twice as tall than right-handed individuals'. Formally, this is a subset of the original claim and is therefore a valid example of specialisation as we define it. However, the extra constraint of left-handed individuals being `more than twice as tall' as right-handed individuals is so blatantly improbable that a system capable of any degree of moderately well-calibrated uncertainty quantification would assign a near-zero probability to it, thus undermining the informativeness of the task. 

Therefore, to ensure the generated examples conform to these expectations, and fulfil the criteria for each of the respective properties, we again manually check each generated sample. We adjust any samples that were found to violate either of these requirements.

\section{Task Description}
\label{sec:tasks}

We carry out experiments to illustrate the degree to which LLMs adhere to the principles of complementarity and monotonicity. For the former, we compare the base claim with a `specialised' claim which is a subset of the base claim, and a `generalised' claim which is a superset. For the latter,  we compare the models' uncertainty in a `base' claim, with the uncertainty in the negated version of the claim (i.e. the complement of the claim). Following is a description of the tasks we use for evaluation.

\begin{table*}[ht]
\centering
\sisetup{output-decimal-marker={.}, group-digits=false}
\begin{tabular}{@{} c 
                | *{2}{S[table-format=1.3]@{\hspace{12pt}}} 
                | *{2}{S[table-format=1.3]@{\hspace{12pt}}} 
                | *{2}{S[table-format=1.3]@{\hspace{12pt}}} 
                @{}}
\toprule
\multirow{2}{*}{\centering\textbf{Model}} & 
\multicolumn{2}{c|}{\textbf{Direct Prompting}} & 
\multicolumn{2}{c|}{\textbf{Chain-of-Thought}} & 
\multicolumn{2}{c}{\textbf{Argumentative LLM}} \\
\cmidrule(lr){2-3} \cmidrule(lr){4-5} \cmidrule(lr){6-7}
 & \multicolumn{1}{c}{Spec.} & \multicolumn{1}{c|}{Gen.} & 
 \multicolumn{1}{c}{Spec.} & \multicolumn{1}{c|}{Gen.} & 
 \multicolumn{1}{c}{Spec.} & \multicolumn{1}{c}{Gen.} \\
\midrule
Llama-3        & 0.595 & 0.725 & 0.565 & 0.712 & 0.582 & 0.750 \\
Llama-3.1      & 0.590 & 0.748 & 0.615 & 0.785 & 0.598 & 0.745 \\
Mixtral        & 0.548 & 0.738 & 0.515 & 0.730 & 0.514 & 0.655 \\
Gemma          & 0.452 & 0.515 & 0.462 & 0.545 & 0.500 & 0.507 \\
Gemma-2        & 0.618 & 0.788 & 0.598 & 0.775 & 0.550 & 0.770 \\
Gpt-4o-mini    & 0.627 & 0.771 & 0.610 & 0.767 & \textbf{0.692} & \textbf{0.835} \\
Gpt-4o*        & 0.660 & 0.740 & 0.660 & 0.790 & 0.620 & 0.820 \\
\midrule
\multicolumn{1}{l|}{\textbf{Average}} & 
\textbf{0.584} & 0.718 & 0.576 & \textbf{0.729} & 0.582 & 0.726 \\
\bottomrule
\end{tabular}
\caption{Accuracy for the \textbf{monotonicity} task across all evaluated models for two sets of 400 claim pairs (1200 samples total). Values represent performance on \textbf{specialisation} (Spec.) and \textbf{generalisation} (Gen.) tasks. Each column refers to different uncertainty quantification techniques described in Section \ref{sec: uncertainty}. Best performances highlighted in bold.}
\label{tab:monontonicity}
\vspace{0.2cm}
\small
\begin{tabular}{@{}p{0.9\textwidth}@{}}
*Reduced sample size evaluation (100 claim pairs per task).
\end{tabular}
\end{table*}

\begin{table*}[ht]
\centering
\sisetup{output-decimal-marker={.}, group-digits=false}
\begin{tabular}{@{} c 
                | *{3}{S[table-format=1.3]@{\hspace{12pt}}} 
                | *{3}{S[table-format=1.3]@{\hspace{12pt}}} 
                | *{3}{S[table-format=1.3]@{\hspace{12pt}}} 
                @{}}
\toprule
\multirow{2}{*}{\centering\textbf{Model}} & 
\multicolumn{3}{c|}{\textbf{Direct Prompting}} & 
\multicolumn{3}{c|}{\textbf{Chain-of-Thought}} & 
\multicolumn{3}{c}{\textbf{Argumentative LLM}} \\
\cmidrule(lr){2-4} \cmidrule(lr){5-7} \cmidrule(lr){8-10}
 & \multicolumn{1}{c}{$>$5\%} & \multicolumn{1}{c}{$>$10\%} & \multicolumn{1}{c|}{$>$15\%} & 
 \multicolumn{1}{c}{$>$5\%} & \multicolumn{1}{c}{$>$10\%} & \multicolumn{1}{c|}{$>$15\%} & 
 \multicolumn{1}{c}{$>$5\%} & \multicolumn{1}{c}{$>$10\%} & \multicolumn{1}{c}{$>$15\%} \\
\midrule
Llama-3        & 0.965 & 0.942 & 0.905 & 0.960 & 0.932 & 0.892 & 0.858 & 0.720 & 0.615 \\
Llama-3.1      & 0.850 & 0.765 & 0.675 & 0.908 & 0.822 & 0.740 & 0.840 & 0.735 & 0.652 \\
Mixtral        & 0.870 & 0.775 & 0.700 & 0.892 & 0.788 & 0.728 & 0.890 & 0.775 & 0.719 \\
Gemma          & 0.818 & 0.662 & 0.502 & 0.822 & 0.623 & 0.468 & 0.922 & 0.842 & 0.748 \\
Gemma-2        & 0.922 & 0.832 & 0.788 & 0.922 & 0.845 & 0.775 & 0.850 & 0.720 & 0.613 \\
Gpt-4o-mini    & 0.766 & \textbf{0.529} & 0.446 & 0.739 & \textbf{0.529} & \textbf{0.441} & 0.770 & 0.572 & 0.458 \\
Gpt-4o*        & 0.820 & 0.590 & 0.460 & 0.800 & 0.560 & 0.460 & \textbf{0.720} & 0.610 & 0.510 \\
\midrule
\multicolumn{1}{l|}{\textbf{Average}} & 
0.856 & 0.721 & 0.640 & 0.863 & 0.727 & 0.646 & \textbf{0.833} & \textbf{0.711} & \textbf{0.599} \\
\bottomrule
\end{tabular}
\caption{Performance on the \textbf{complementarity} task across all evaluated models for two sets of 400 claim pairs (1200 samples total). Values represent the proportion of predictions deviating by more than 5\%, 10\%, and 15\% from ground truth confidence scores (i.e. $|$1 - \textit{P(original statement) + P(negated statement)}$|$ * 100). Bold values indicate best performance per threshold column; note the smallest amount of deviation indicates the best performance.}
\label{tab:deviation}
\vspace{0.2cm}
\small
\begin{tabular}{@{}p{0.9\textwidth}@{}}
*Reduced sample size evaluation (100 claim pairs). 
\end{tabular}
\end{table*}

\subsection{Monotonicity}
\label{sec:monot}

For measuring a model's ability to demonstrate probabilistic beliefs that adhere to the property of monotonicity, we compare the base claim with the specialised and generalised version of the claim (as described in Section \ref{sec:dataset}) respectively. Therefore, to assess conformity to monotonicity, we measure whether the following inequalities hold:

\textit{P(original statement) $>$ P(specialised statement)}.

\textit{P(original statement) $<$ P(generalised statement)}.

We note that we define the inequalities as strict in both cases. While, theoretically, one can generate more/less specific versions of a claim that remain equally as likely, the practical instantiation of our dataset means that the `difference' between the modified and original claim are always non-zero. 

For instance, once again referring to the example in Figure \ref{fig:examples}, the difference between the original claim and the generalisation is something like: `In a typical human population, left-handed individuals have the same IQ as right-handed individuals'. This statement clearly has a greater than zero probability, meaning that the conjunction of it with the original claim must result in an increased probability.

\subsection{Complementarity}
\label{sec:compl}

In order to evaluate adherence to complementarity, we assess the deviation of a model's evaluation of a claim's uncertainty, with that of the negated version of the same claim. Then, complementarity amounts to obtaining adherence to the following equation: 

\textit{P(original statement) + P(negated statement)} = 1.

A deviation from the value of 1 (i.e. either greater or less than) is considered a failure of the model to exhibit rational probabilistic beliefs, permitting some room for noise below a reasonable threshold.

\section{Uncertainty Quantification}
\label{sec: uncertainty}

In order to quantify the models' performance on the tasks outlined in Section \ref{sec:tasks}, we use three different methods to elicit probability estimates. We also experiment with a fourth for the larger, closed-source models only, due to technical constraints we discuss in more detail in the relevant section. Below, we describe each approach.

\subsection{Direct Prompting}
\label{sec: direct}

The first uncertainty quantification method we assess is a simple prompting based approach, wherein a prompt, including the target claim, is input into a LLM and the probability is taken directly from the generated output. Due to the high variation in performance depending on the formulation of the prompt \cite{zhuo2024prosa}, we form a validation set from 468 samples (117 of each variety of claim) in our dataset. Using this validation set we assess the performance of four distinct prompt variations.

We take a principled approach to candidate prompt selection. The first candidate is taken directly from existing literature \cite{freedman2024argumentativelargelanguagemodels}. The other three candidates were generated by GPT-o1 (using the web interface), having been provided a detailed task description. The best performing candidate was the one taken from the literature:

\begin{tcolorbox}[colframe=blue, colback=blue!5!white, sharp corners, boxrule=0.5mm, width=\columnwidth]
\small
\textbf{You are an analyst evaluating the validity of statements.}

For the statement:

\smallskip

\textbf{Statement:} \texttt{"\{statement\}"}

\smallskip

Please give your confidence that the statement is correct, accurate, and truthful. Your response should be between \textbf{0\% and 100\%}, with \textbf{0\%} indicating the statement is definitely invalid, \textbf{100\%} indicating it is definitely valid, and values in between indicating varying levels of uncertainty. 

Your estimates should be well-calibrated, so feel free to err on the side of caution and output moderate probabilities if you are not completely sure in your assessment. 

\textbf{Only respond with your confidence score, no other text.}

Please respond in the following form:

\smallskip

\textbf{Likelihood:} The predicted likelihood that the considered statement is valid

\end{tcolorbox}

When running the experiments the `statement' variable is replaced by the target claim being evaluated. The probability estimate is extracted from the generated output using simple pattern matching. In order to negate the sampling bias introduced by the indeterminacy of LLM outputs, we generate 5 outputs for each sample, and take the mean of the scores as the final probability.

\subsection{Chain-of-Thought}

The chain-of-thought \cite{chain-of-thought2022} approach we adopt is essentially a simple extension of the direct prompting method. While some instantiations of chain-of-thought use few-shot examples of how a problem can be broken down into discrete steps, this is more applicable to formal settings such as for mathematical problems. Instead, we simply append the sentence: `Let's think step by step', to the end of the prompt, as introduced in \citeauthor{10.5555/3600270.3601883} \citeyear{10.5555/3600270.3601883}.

Similarly to the direct prompting approach, we use the mean of 5 outputs as the final result.
 
\subsection{Argumentative Large Language Models}
\label{sec: allms}

Our third approach is taken from the literature on claim verification. \citeauthor{freedman2024argumentativelargelanguagemodels} (\citeyear{freedman2024argumentativelargelanguagemodels}) develop a method that uses outputs from LLMs to create formal arguments, called argumentative LLMs (ArgLLMs). Each argument is assigned a confidence score, also by LLMs, and then, using reasoning techniques from the Computational Argumentation literature, a final score is computed. 

We follow 
a specific instantiation of ArgLLMs outlined in 
\citeauthor{freedman2024argumentativelargelanguagemodels} (\citeyear{freedman2024argumentativelargelanguagemodels}). This involves generating one argument supporting and one attacking the `topic' claim, and then an additional layer of supporting and attacking arguments for both of these arguments. An `argument strength' is also generated for each of the arguments, as well as the topic claim, meaning that six arguments and seven argument strengths are output for each claim. 

For more specific details of the system, including the argument generation, argument strength attribution and final uncertainty calculation processes, please refer to 
\citeauthor{freedman2024argumentativelargelanguagemodels} (\citeyear{freedman2024argumentativelargelanguagemodels}). Due to the large number of generations required for for each sample, and our computational constraints, we are limited to producing one output per claim. 

\subsection{Top-K Logit Sampling}
\label{sec:topk}

\begin{table*}[ht]
\centering
\sisetup{output-decimal-marker={.}, group-digits=false}
\begin{tabular}{@{} c 
                | c
                | c
                | *{3}{S[table-format=1.3]@{\hspace{8pt}}} 
                @{}}
\toprule
\multirow{2}{*}{\textbf{Model}} &
\multirow{2}{*}{\textbf{Specification}} &
\multirow{2}{*}{\textbf{Generalisation}} &
\multicolumn{3}{c}{\textbf{Complementarity}} \\
\cmidrule(lr){4-6}
 &  &  & \multicolumn{1}{c}{$>$5\%} & \multicolumn{1}{c}{$>$10\%} & \multicolumn{1}{c}{$>$15\%} \\
\midrule
GPT-4o-mini & 0.712 & 0.825 & \textbf{0.729} & \textbf{0.549} & \textbf{0.441} \\
GPT-4o*      & \textbf{0.760} & \textbf{0.830} & 0.750 & 0.580 & 0.520 \\
\bottomrule
\end{tabular}
\caption{Results for Top-K Logit Sampling method described in Section \ref{sec:topk}.}
\label{tab:topk}
\vspace{0.2cm}
\small
\begin{tabular}{@{}p{0.9\textwidth}@{}}
*Reduced sample size evaluation (100 claim pairs). 
\end{tabular}
\end{table*}

For our final method, we leverage the ability to obtain raw logit outputs from the models. These can be turned into a probability distribution over the entire token space by applying a softmax to the logits. In our case we limit this to the top 5 most likely next tokens. Since the tokens themselves are numerical (we use the same prompt as in Section~\ref{sec: direct}), we are able to calculate a weighted average of the token value by the probability distribution.

Formally, let \(\mathcal{V}\) denote the full vocabulary, comprising all possible tokens. For each token \(i \in \mathcal{V}\), let \(z_i\) be the raw logit, and define the softmax probability:
\[
p_i = \frac{\exp\bigl(z_i\bigr)}{\sum_{j=1}^{|\mathcal{V}|} \exp\bigl(z_j\bigr)}.
\]
Let \(S \subseteq \mathcal{V}\) be the set of \(k\) tokens with the highest probabilities \(p_i\). Suppose each token in \(S\) corresponds to a numerical value \(t_i\). Then the final numeric output is:
\[
\hat{y} = \sum_{i \in S} p_i \, t_i.
\]

For instance an example output is:

\[
\begin{aligned}
    &\text{$t_1$: } 95 \quad \text{$p_1$: } 0.803 \\
    &\text{$t_2$: } 90 \quad \text{$p_2$: } 0.179 \\
    &\text{$t_3$: } 98 \quad \text{$p_3$: } 0.010 \\
    &\text{$t_4$: } 85 \quad \text{$p_4$: } 0.007 \\
    &\text{$t_5$: } 80 \quad \text{$p_5$: } 0.001
\end{aligned}
\]

The weighted average of these outputs is then calculated as 94.

A shortcoming of this method is that for the smaller, open-source models we experiment with, we were not able to effectively apply this method. This is because they did not consistently conform to the formatting instructions requiring them to only return numerical outputs. Thus, the models did not return exclusively numerical token candidates, invalidating the weighted average calculation. We give a full breakdown of all models we experiment with in the next section.

\section{Experiments}
\label{sec:exp}

\begin{figure*}[htbp]
\centering
\includegraphics[scale=0.5]{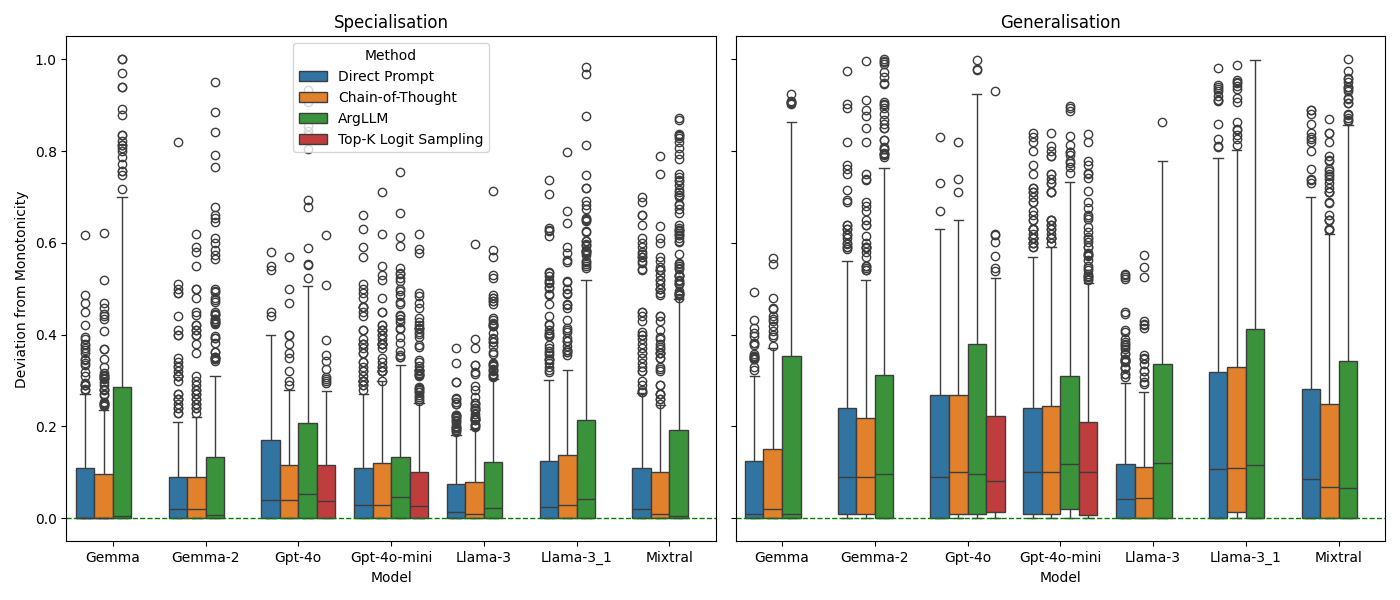}
\caption{Adherence to \textbf{monotonicity} by model and uncertainty quantification methodology. The left panel is the Specialisation task, and the right Generalisation. Both tasks are described in detail in Section~\ref{sec:monot}. The y-axis represents the magnitude of deviation from correctly monotonic probability estimations, and the x-axis is the model type.}
\label{fig:monotonicity}
\end{figure*}

\begin{figure}[htbp]
\centering
\includegraphics[scale=0.32]{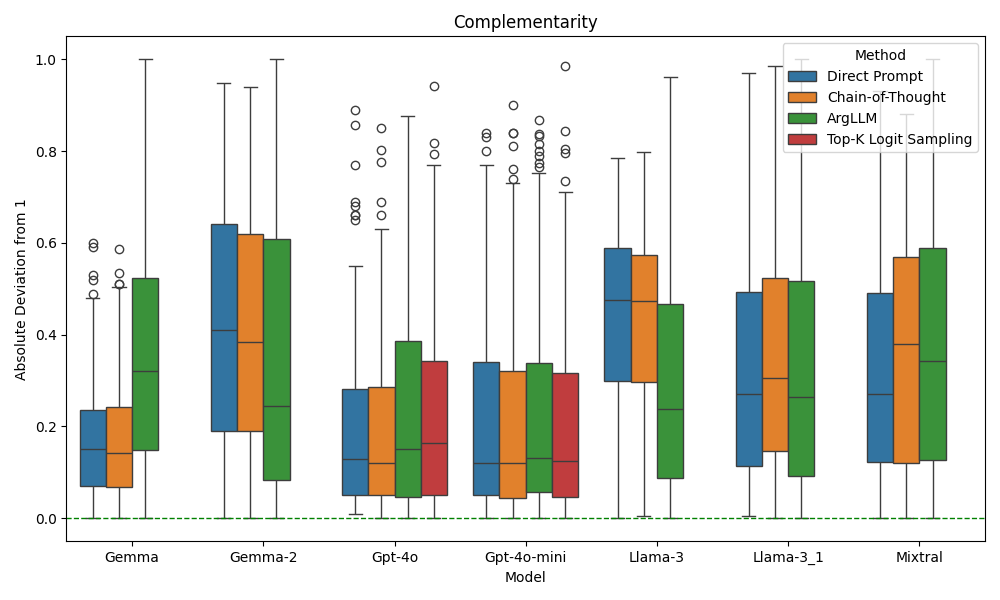}
\caption{Adherence to \textbf{complementarity} by model and uncertainty quantification methodology. Detailed task description is provided in Section~\ref{sec:compl}. 
}
\label{fig:complementarity}
\end{figure}

In this section we provide details about the experimental set-up, as well as a discussion of the results.

\subsection{Models}

We use a number of open and closed-source models for our experiments, to ensure our findings are robust across different model sizes and architectures. We use seven main models: Mixtral (Mixtral-8x7B-Instruct-v0.1)~\cite{jiang2024mixtral}, Gemma (gemma-7b-it)~\cite{gemmateam2024gemma}, Gemma 2 (gemma-2-9b-it)~\cite{gemma2}, Llama 3 (Meta-Llama-3-8B-Instruct)~\cite{llama3}, LLama 3.1 (Meta-Llama-3.1-8B-Instruct), GPT-4o mini (gpt-4o-mini-2024-07-18)~\cite{gpt-4o-mini} and GPT-4o (gpt-4o-2024-08-06)~\cite{gpt-4o}.

For the former six models we use the entire test set of 400 claims and their three variations each (1600 samples in total), and for GPT-4o, due to the cost associated with using closed-source models, we use a subset of 100 claims and their variations (400 samples in total). The model hyperparameters we use are the same across all models: temperature 0.7, top-p 0.95 and repetition penalty 1.0. In order to reduce the computational cost of running the open-source models we quantize all of them to 4-bits \cite{dettmers2024qlora}.

\subsection{Results}

The results for all the main experiments, excluding prompt selection, are reported in Tables \ref{tab:monontonicity} and \ref{tab:deviation}. Additional results for the Top-K Logit Sampling method (Section \ref{sec:topk}), are reported for a reduced set of models in Table \ref{tab:topk}. Results are represented graphically in Figures \ref{fig:monotonicity} and \ref{fig:complementarity}.

\paragraph{Discussion}

Both the results reported in Table~\ref{tab:monontonicity} and Table~\ref{tab:deviation} suggest that the larger, closed-source models (specifically GPT-4o-mini and GPT-4o) perform better at both tasks. This appears to confirm the well-established relationship between model size and performance \cite{liang2022holistic}.

Despite the consistently stronger performance of the larger models, neither GPT-4o-mini nor GPT-4o showed robust adherence to all of the tested properties. We still observe that both can produce contradictory or incoherent outputs in terms of basic probability axioms. In particular, the violation rates for complementarity, while lower than those of other models, are still non-trivial. This suggests that even state-of-the-art LLMs, although generally more capable, may not maintain fully consistent probabilistic representations across diverse claim types and domains.

Examining the different uncertainty quantification techniques reveals additional insights. Chain-of-thought prompting offers some improvement over direct prompting, especially for smaller models, as it encourages more explicit reasoning steps that may guide the model to produce more stable probability assessments. However, for larger models, we find only marginal gains with chain-of-thought. 

The ArgLLM technique achieves similar or slightly better performance than chain-of-thought in some scenarios, particularly for closed-source systems. By introducing additional structure via conflicting viewpoints and supporting/attacking sub-arguments, this approach may force the model to reconcile its own inconsistencies to some degree. However, it remains computationally expensive, as it entails generating multiple layers of arguments and confidence assignments for every claim. 

The Top-K Logit Sampling method, examined only for GPT-4o-mini and GPT-4o in our experiments, showed modest improvements in monotonicity for GPT-4o, but not for GPT-4o-mini. While extracting probabilities at the token level mitigates some of the uncertainty introduced by sampling, it is limited by whether the model consistently follows the numerical response format. Our attempts to replicate this approach with the open-source models were largely unsuccessful, demonstrating the fragility of logit-based methods when generation style cannot be strictly controlled.

Moreover, when we analyse Figure \ref{fig:monotonicity}, we see that despite having a smaller proportion of monotonically unsound probability estimations, when the larger models are incorrect they are so by a greater magnitude. This can seen clearly, for example, by comparing the mean deviation on the generalisation task for GPT-4o and GPT-4o-mini, with that of Gemma, which is lower across all methods. Interestingly, Figure \ref{fig:complementarity}, while demonstrating the generally superior performance of the larger models, also appears to show Gemma as performing almost as well as the larger models, when using the direct prompting and chain-of-thought methods.

\section{Related Work}
\label{sec:related}

Our findings build on a body of work demonstrating weaknesses in the ability of LLMs to adhere to basic logical principles. The reversal curse \cite{berglund2023reversal} shows that in cases where a model has learnt ``A is B'', it has often not learnt ``B is A''. We demonstrate a similar effect but with probabilistic beliefs. 

Similarly, \citeauthor{fluri2023evaluating} (\citeyear{fluri2023evaluating}) demonstrate that 
LLMs succumb to various logical inconsistencies. They also show that LLMs can exhibit non-monotonic estimates, but restrict their analysis to the task of forecasting.

\citeauthor{Wong2023FromWM} (\citeyear{Wong2023FromWM}) propose a method for integrating LLMs with a probabilistic logic engine. They also argue the necessity for LLMs to be able to reason probabilistically, and suggest that combining them with a symbolic module is the best way to achieve this.
%

\citeauthor{kuhn2022semantic} (\citeyear{kuhn2022semantic}) also note the insufficiency of using direct prompting-based methods to ascertain the uncertainty of LLM outputs, instead developing a sampling-based method. 
However, this technique addresses subjective model uncertainty, and it is not clear whether there is an effective method to adapt it for representing objective uncertainty.

\section{Conclusion and Future Work}

In this paper we demonstrate that state-of-the-art LLMs fail at basic probabilistic reasoning. We observe that larger models demonstrate improved performance relative to their smaller counterparts. Nevertheless, the current scale of their failure is too significant to assume that further scaling will completely eradicate the problem.


A neurosymbolic approach, such as the one presented in \citeauthor{Wong2023FromWM} (\citeyear{Wong2023FromWM}), bypasses the need for LLMs to reason probabilistically. Instead, they may rely on symbolic modules to handle any probabilistic inferences they may need to make. It is possible that a similar approach, which uses an appropriate symbolic knowledge representation, can provide an effective and robust solution to the problems we highlight in this paper.


\section*{Acknowledgments}

This research was partially supported by ERC under the EU's Horizon 2020 research and innovation programme (grant agreement No. 101020934, ADIX), by J.P.
Morgan and the Royal Academy of Engineering under the Research Chairs and Senior Research Fellowships
scheme, and by UKRI through the CDT in Safe
and Trusted Artificial Intelligence (Grant No. EP/S023356/1).






\bibliographystyle{flairs}
\bibliography{related}

\end{document}